\documentclass{article}

\usepackage{arxiv}

\usepackage[utf8]{inputenc}
\usepackage[T1]{fontenc}
\usepackage{microtype}
\usepackage{hyperref}
\usepackage{url}

\usepackage{amsmath,amssymb,amsfonts}
\usepackage{booktabs}
\usepackage{nicefrac}
\usepackage{graphicx}
\usepackage{subcaption}
\usepackage{caption}
\usepackage{algorithm}
\usepackage{algpseudocode}
\usepackage{comment}
\usepackage{colortbl}
\usepackage[table]{xcolor}
\usepackage{tikz}
\usepackage{multicol}
\usepackage{natbib}
\usepackage{listings}
\usepackage{longtable}
\usepackage{adjustbox}
\usepackage{array}
\usepackage{tcolorbox}
\tcbuselibrary{skins}

\usetikzlibrary{
  positioning,
  arrows.meta,
  shapes.geometric,
  shapes.multipart,
  fit,
  calc,
  backgrounds
}

\graphicspath{{./}}

\newcommand{\smalltag}{\footnotesize\color{black!70}}

\tcbset{
  enhanced,
  boxrule=0.5pt,
  arc=2pt,
  left=6pt,
  right=6pt,
  top=5pt,
  bottom=5pt
}

\title{Understanding Pruning Regimes in Vision-Language Models through Domain-Aware Layer Selection}

\author{
 Saeed Khaki \\
  Microsoft AI\\
  \texttt{saeedkhaki@microsoft.com} \\
   \And
 Nima Safaei \\
  Ohio State University\\
  \texttt{safaei.3@osu.edu} \\
  \And
 Kamal Ginotra \\
  Microsoft AI\\
  \texttt{kamalginotra@microsoft.com} \\
}

\begin{document}
\maketitle
\begin{abstract}
Transformer-based vision-language models (VLMs) contain substantial depth redundancy, yet the effect of removing specific decoder layers remains poorly understood, especially for domains that require tight coupling between perception and multi-step reasoning. We study structured decoder layer pruning through the lens of \emph{domain-aware activation similarity}, measuring how strongly each layer transforms representations for math versus non-math inputs. This yields simple math-aware, non-math-aware, and mixed ranking criteria that identify layers whose input--output activations change least within a target domain. Across two state-of-the-art VLMs and a broad suite of math and general multimodal benchmarks, we uncover a consistent three-regime structure: at low pruning budgets, performance is highly sensitive to which layers are removed; at moderate budgets, methods converge as structural damage accumulates; and at high budgets, structural continuity dominates, favoring spacing-aware strategies. Our domain-aware rankings achieve the strongest stability in the ranking-sensitive regime, while matching or exceeding structure-aware baselines at larger budgets. These results provide a clearer picture of how depth contributes to domain-specific behavior in VLMs and offer a practical, interpretable approach to reducing model depth without sacrificing essential mathematical or general vision-language capabilities.
\end{abstract}


\section{Introduction}
Vision-language models (VLMs) have become a practical foundation for multimodal assistants, combining strong visual perception with open-ended language generation and instruction following.
Modern systems typically pair a visual encoder with an autoregressive transformer decoder and are trained with large-scale multimodal instruction data, enabling broad competence in captioning, visual question answering, document understanding, and multimodal reasoning \citep{vaswani2017attention,li2023blip2,liu2023llava,li2024llavaonevision,qwen3_vl_tech_report}.
This scaling trend has also improved performance on demanding benchmarks that require multi-step reasoning over images, including visual math and STEM-style problems \citep{lu2022scienceqa,qwen3_vl_tech_report}.
At the same time, the resulting models are increasingly expensive to deploy, motivating compression techniques that reduce latency and memory while preserving critical capabilities.

Pruning is a particularly attractive direction because it can produce hardware-friendly speedups when it removes structured components such as attention heads or whole transformer blocks.
A line of work has shown that substantial redundancy exists in transformer architectures, including attention heads that can often be removed with limited impact \citep{michel2019sixteen,voita2019heads}.
Other methods target depth directly, for example by training models that can operate at reduced depth or by selecting and removing layers after training \citep{fan2019layerdrop,peer2022greedy_layer}.
More recent efforts study post-training sparsification and structured compression for large language models, including one-shot pruning and activation-aware criteria \citep{frantar2023sparsegpt,sun2024wanda}, as well as structural approaches that prune coupled components and recover performance via lightweight adaptation \citep{ma2023llmpruner,ashkboos2024slicegpt}.
Despite rapid progress, most pruning strategies are evaluated using aggregate quality metrics, leaving open a central question for multimodal deployment: which parts of a VLM are essential for a target domain, and when does the exact choice of removed layers matter?

This question is especially acute for mathematical reasoning in VLMs.
Compared to generic captioning or object-centric VQA, math problems in the wild frequently require a tight coupling of perception (symbols, diagrams, tables, layout) and multi-step language reasoning.
Small structural changes can therefore cause disproportionate failures, even when average vision-language scores appear stable.
In language-only models, pruning has been observed to exhibit distinct regimes as sparsity increases, where mild pruning is nearly free, intermediate pruning induces transfer failures, and aggressive pruning becomes capacity-limited \citep{gordon2020compressing}.
Whether an analogous regime structure exists for depth pruning in VLMs, and how it interacts with domain specialization, is not well understood.

In this work, we study \emph{structured decoder layer pruning} for VLMs with the explicit goal of preserving mathematical reasoning while maintaining general vision-language competence.
Our key idea is simple: a layer that barely changes its hidden representation for inputs from a particular domain is likely redundant for that domain.
We operationalize this by logging the input and output activations of each decoder block on a diverse set of domain-specific probes and measuring input--output cosine similarity.
This yields two complementary redundancy signals, one computed on math-oriented prompts and one computed on non-math prompts.
Using these signals, we rank layers with a domain-aware criterion and prune under a fixed budget while preserving endpoint layers to avoid destabilizing early multimodal fusion and final token prediction.
After pruning, we apply a brief supervised fine-tuning stage to restore stability, using the same adaptation protocol for all methods to ensure a fair comparison.

We evaluate two state-of-the-art VLMs, Qwen3-VL-2B-Instruct and Qwen3-VL-4B-Instruct \citep{qwen3_vl_tech_report}, and report results on both math-focused benchmarks (Snapask, NuminaMath-CoT) and general-purpose benchmarks covering chart understanding, real-world spatial reasoning, multimodal science QA, and fine-grained perception \citep{masry2022chartqa,realworldqa_xai,lu2022scienceqa,wu2024vstar,zhang2024lmms_eval}.
Our experiments reveal a consistent \emph{three-regime} pattern in decoder-depth pruning. At low pruning budgets, performance is highly sensitive to which layers are removed, making domain-aware ranking the most effective strategy.

At moderate budgets, methods begin to converge as structural damage accumulates.
At high budgets, structural continuity becomes the limiting factor, and strategies that enforce spacing between deletions can dominate, aligning with recent structure-aware layer pruning for VLMs \citep{Madinei2025INTERLACE}.
Across regimes, random deletion is the most brittle baseline, highlighting that redundancy in transformers is real but not uniformly distributed across depth.

\paragraph{Contributions.}
\begin{itemize}
  \item We introduce a domain-aware decoder layer pruning method for VLMs based on input--output activation similarity, yielding math-aware, non-math-aware, and mixed ranking variants.
  \item We provide an empirical regime analysis of depth pruning in VLMs, identifying ranking-sensitive, transition, and structure-dominated regimes that explain when domain-aware selection matters.
  \item We benchmark against representative baselines, including representation-similarity pruning and structure-aware layer pruning, and report results on both math and general vision-language evaluations.
\end{itemize}

\section{Related Work}

Pruning in Transformer models has been explored through head removal, structured depth reduction, and magnitude-based sparsification, demonstrating that many components contribute unevenly to model performance \cite{michel2019sixteen, voita2019heads, fan2019layerdrop, frantar2023sparsegpt, sun2024wanda, ma2023llmpruner, ashkboos2024slicegpt}. These approaches, however, are primarily developed for text-only LLMs and do not account for multimodal activation patterns.

Vision-language pruning has recently gained attention. INTERLACE introduced an interleaved prune--adapt strategy for VLMs, showing that local redundancy can be exploited effectively \citep{Madinei2025INTERLACE}. Other work has examined modality-specific pruning sensitivity, revealing asymmetries between visual and textual representations \citep{pmlr-v97-kornblith19a}. These methods remain modality-agnostic and do not incorporate domain-specific activation signals.

Representation similarity metrics such as CKA provide a complementary view of redundancy by identifying layers with highly aligned activation structures \cite{pmlr-v97-kornblith19a}. While CKA has been used for analysis, it has not been applied in a domain-conditioned manner for pruning. Our method differs by leveraging task-specific activation patterns to guide structured pruning tailored to mathematical reasoning.

\section{Method}
\label{sec:method}
We study structured layer pruning for vision-language models (VLMs) with the goal of preserving mathematical reasoning while maintaining general vision-language performance. Our approach ranks transformer decoder layers by domain-specific activation similarity---measuring how much each layer transforms representations for math versus non-math inputs---and removes layers deemed redundant for a target domain. After pruning, we apply a lightweight post-pruning supervised fine-tuning (SFT) to recover capacity lost due to structural changes.
We consider three variants of domain-aware ranking: math-aware, non-math-aware, and mixed. We then compare them against structure-agnostic and structure-aware baselines. Our method is intentionally simple, interpretable, and composable with existing pruning strategies.

\subsection{Problem Setup}
\label{sec:setup}
We consider a vision-language model based on the transformer architecture
\cite{vaswani2017attention}, following recent designs that combine a visual
encoder, a projection module for cross-modal alignment, and an autoregressive
language decoder \cite{li2023blip2,liu2023llava}.
Given an image \( \mathcal{I} \) and a text prompt \( \mathcal{P} \), the visual
encoder produces visual tokens
\begin{equation}
\mathbf{Z}_v = \phi(\mathcal{I}) \in \mathbb{R}^{N_v \times d_v},
\end{equation}
which are projected into the language embedding space as
\begin{equation}
\tilde{\mathbf{Z}}_v = \psi(\mathbf{Z}_v) \in \mathbb{R}^{N_v \times d_h}.
\end{equation}
The prompt is embedded as
\begin{equation}
\mathbf{Z}_t = \mathrm{Emb}(\mathcal{P}) \in \mathbb{R}^{N_t \times d_h},
\end{equation}
and the decoder input is formed by concatenation:
\begin{equation}
\mathbf{H}^{(0)} = [ \tilde{\mathbf{Z}}_v ; \mathbf{Z}_t ].
\end{equation}

The language decoder consists of \( L \) transformer layers producing hidden
states
\begin{equation}
\mathbf{H}^{(\ell)} = \mathcal{D}_{\ell}(\mathbf{H}^{(\ell-1)}).
\end{equation}
Here \(N_v\) and \(N_t\) denote the number of visual and textual tokens,
\(d_v\) is the visual feature dimension, and \(d_h\) is the hidden size of the
language embedding space.

We adopt depth-wise pruning, in which entire decoder layers
(i.e., transformer blocks) are removed.
The vision encoder and modality projection module are kept fixed,
as prior studies indicate that these components are essential for
maintaining stable cross-modal alignment~\cite{li2023blip2, liu2023llava}.
In addition, we preserve the first and last decoder layers,
since they play a critical role in multimodal feature integration
and final token prediction.

Let \( \mathcal{L}_{\mathrm{mid}} \) denote the set of pruneable decoder layers.
Given a pruning budget \( p \in \{10\%, 25\%, 40\%\} \), we remove
\( K = \lfloor p \cdot |\mathcal{L}_{\mathrm{mid}}| \rfloor \) layers.


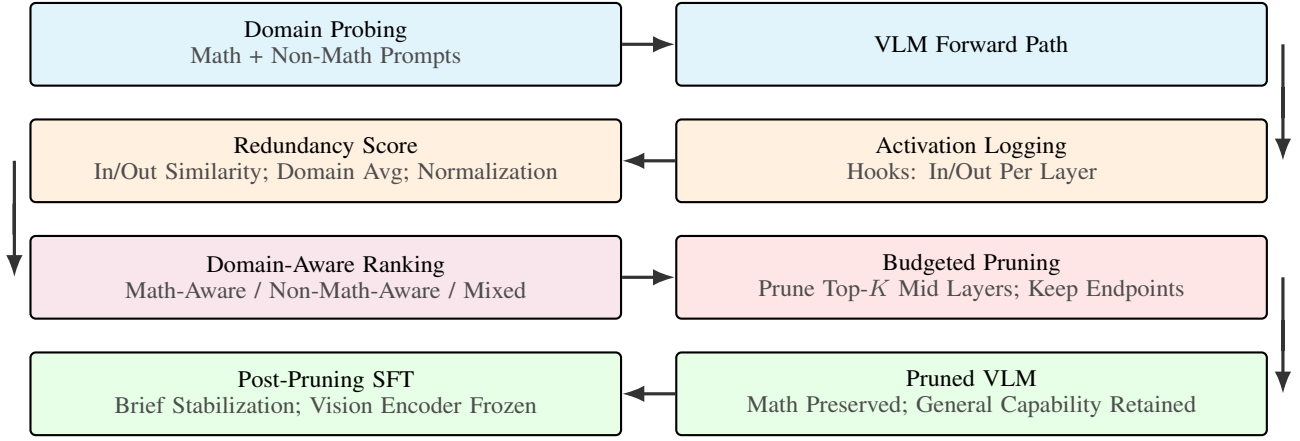
\begin{figure}[t]
\centering
\begin{tikzpicture}[
  font=\small,
  >=Latex,
  arr/.style={->, very thick, draw=black!80},
  box/.style={draw, rounded corners=2pt, thick, align=center,
              inner sep=6pt, minimum height=11mm, text width=0.45\linewidth},
  A/.style={box, fill=cyan!10},
  B/.style={box, fill=orange!12},
  C/.style={box, fill=purple!10},
  D/.style={box, fill=red!10},
  E/.style={box, fill=green!10},
  wrap/.style={arr, rounded corners=8pt}
]

\node[inner sep=0pt, outer sep=0pt, minimum width=\linewidth] (canvas) {};
\coordinate (W) at (canvas.west);

\node[A, anchor=west] (probe) at ($(W)+(0,0)$)
{Domain Probing\\\smalltag Math + Non-Math Prompts};

\node[A, anchor=west] (fwd) at ($(W)+(0.52\linewidth,0)$)
{VLM Forward Path};

\draw[arr] (probe.east) -- (fwd.west);

\node[B, anchor=west] (log) at ($(W)+(0.52\linewidth,-1.55)$)
{Activation Logging\\\smalltag Hooks: In/Out Per Layer};

\node[B, anchor=west] (score) at ($(W)+(0,-1.55)$)
{Redundancy Score\\\smalltag In/Out Similarity; Domain Avg; Normalization};

\draw[arr] (log.west) -- (score.east);

\coordinate (r1end) at ($(fwd.east)+(2mm,0)$);
\coordinate (r2start) at ($(log.east)+(2mm,0)$);
\draw[wrap] (r1end) -- ($(r1end)+(0,-7mm)$) -- ($(r2start)+(0,7mm)$) -- (r2start);

\node[C, anchor=west] (rank) at ($(W)+(0,-3.10)$)
{Domain-Aware Ranking\\\smalltag Math-Aware / Non-Math-Aware / Mixed};

\node[D, anchor=west] (prune) at ($(W)+(0.52\linewidth,-3.10)$)
{Budgeted Pruning\\\smalltag Prune Top-$K$ Mid Layers; Keep Endpoints};

\draw[arr] (rank.east) -- (prune.west);

\coordinate (l2end) at ($(score.west)+(-2mm,0)$);
\coordinate (l3start) at ($(rank.west)+(-2mm,0)$);
\draw[wrap] (l2end) -- ($(l2end)+(0,-7mm)$) -- ($(l3start)+(0,7mm)$) -- (l3start);

\node[E, anchor=west] (final) at ($(W)+(0.52\linewidth,-4.65)$)
{Pruned VLM\\\smalltag Math Preserved; General Capability Retained};

\node[E, anchor=west] (sft) at ($(W)+(0,-4.65)$)
{Post-Pruning SFT\\\smalltag Brief Stabilization; Vision Encoder Frozen};

\draw[arr] (final.west) -- (sft.east);

\coordinate (r3end) at ($(prune.east)+(2mm,0)$);
\coordinate (r4start) at ($(final.east)+(2mm,0)$);
\draw[wrap] (r3end) -- ($(r3end)+(0,-7mm)$) -- ($(r4start)+(0,7mm)$) -- (r4start);

\end{tikzpicture}

\caption{Overview of the domain-aware decoder layer pruning pipeline. Domain-specific prompts drive activation logging; redundancy is estimated from layer in/out similarity and aggregated per domain. Layers are ranked (math-aware, non-math-aware, or mixed) and pruned under a fixed budget with protected endpoints, and a brief SFT stabilizes the pruned model with the vision encoder kept frozen.}
\label{fig:method_overview}
\end{figure}

\subsection{Domain-Specific Activation Capture}
\label{sec:activation_capture}
Our redundancy metric is derived from the similarity between the input and
output hidden representations of each decoder layer during forward inference.
For every decoder layer $\ell$, we capture the hidden states immediately
before and after the layer via framework-level forward hooks.
The input activation
$H^{\mathrm{in}}_{\ell} \in \mathbb{R}^{B \times T \times d}$
denotes the layer input, while the output activation
$H^{\mathrm{out}}_{\ell} \in \mathbb{R}^{B \times T \times d}$
denotes the corresponding layer output,
where $B$ is the batch size, $T$ is the sequence length, and $d$ is the
hidden dimension. Each forward pass is driven by an image-prompt pair presented as a single chat message, with the image followed by text using the model's standard
template and modality tokens. For efficiency and reuse, we mean-pool activations across the batch and token
dimensions to obtain per-layer vectors
\( \bar{h}^{\mathrm{in}}_{\ell} \) and
\( \bar{h}^{\mathrm{out}}_{\ell} \in \mathbb{R}^{d} \),
which are stored in floating-point precision and reused by several baselines. Additional implementation details are provided in Appendix~\ref{sec:activation_details}.

\subsection{Input--Output Similarity as Redundancy}
\label{sec:similarity}

For each forward pass \( i \), we measure the degree to which a decoder layer
transforms its input representation by computing the cosine similarity between
its input and output hidden states at the token level:
\begin{equation}
\mathrm{sim}_{\ell}^{(i)} =
\frac{1}{T}
\sum_{t=1}^{T}
\frac{\langle h^{\mathrm{in}}_{\ell,t}(i),\, h^{\mathrm{out}}_{\ell,t}(i)\rangle}
{\|h^{\mathrm{in}}_{\ell,t}(i)\|\,\|h^{\mathrm{out}}_{\ell,t}(i)\|}.
\end{equation}
A high similarity score indicates that layer \( \ell \) performs only a small
update to the representations for that input, suggesting functional redundancy.

\paragraph{Domain- and subtask-specific probing.}
A key aspect of our method is that each domain is probed using
\emph{multiple carefully designed subtasks} rather than a single generic prompt.
For the math domain, we use Math-CoT, Math-Direct, Math-Rephrase,
Math-Formalize, and Math-Verify prompts; for the non-math domain, we use
captioning, entity listing, counting-style VQA, and referring-expression
grounding prompts. As shown in Sections~\ref{sec:activation_details} and
\ref{sec:prompts_formatted}, each subtask uses a distinct natural-language
instruction paired with the same image.

Using diverse subtasks is essential because different prompt types trigger 
different computation paths inside the decoder: mathematical reasoning,
linguistic paraphrasing, entity-centric grounding, and global scene
understanding each activate distinct attention heads, MLP channels, and
compositional behaviors. Probing with multiple prompts therefore yields a more
comprehensive view of how each layer behaves across domains.

\paragraph{Domain-level aggregation.}
For each domain \( d \in \{\text{math}, \text{nonmath}\} \), we aggregate the
similarity scores across all subtasks and samples:
\begin{equation}
S^{(d)}_{\ell}
=
\frac{1}{N_d}
\sum_{i=1}^{N_d}
\mathrm{sim}_{\ell}^{(i)},
\end{equation}
where the math domain uses \( N_{\text{math}} = 5{,}000 \) forward passes and the
non-math domain uses \( N_{\text{nonmath}} = 4{,}000 \), distributed across their
respective subtasks.

\paragraph{Normalization across layers.}
To make scores comparable across the depth of each model, we apply z-normalization
within each domain:
\begin{equation}
\hat{S}^{(d)}_{\ell}
=
\frac{S^{(d)}_{\ell} - \mu_d}{\sigma_d + \epsilon},
\end{equation}
where \( \mu_d \) and \( \sigma_d \) are the mean and standard deviation computed
over all pruneable layers for domain \( d \).

\subsection{Layer Ranking and Pruning}
\label{sec:ranking}

Layers are ranked according to their redundancy scores and pruned in descending
order under a fixed budget. We consider two domain-specific ranking strategies:
a math-aware strategy using the normalized math scores
\( \hat{S}^{(\text{math})}_{\ell} \), and a non-math-aware strategy using
\( \hat{S}^{(\text{nonmath})}_{\ell} \).
To combine both signals, we also construct a mixed ranking defined as
\begin{equation}
R_{\ell}(\alpha) =
\alpha\,\hat{S}^{(\text{nonmath})}_{\ell}
+ (1-\alpha)\,\hat{S}^{(\text{math})}_{\ell},
\end{equation}
where \( \alpha \in [0,1] \) controls the balance between non-math and math
redundancy.

In practice, we select \( \alpha = 0.7 \), which we found to provide the best
trade-off between maintaining general vision-language ability and preserving
math reasoning performance.
We intentionally give more weight to the non-math domain because core
capabilities such as OCR, VQA, and grounding are essential for stable image
understanding; pruning layers that are functional for these tasks leads to
significant degradation even when math reasoning layers are preserved.
The value \( \alpha = 0.7 \) therefore provides a balanced, empirically robust
ranking that retains mathematical competence without compromising baseline
vision-language behavior.

\subsection{Post-Pruning Stabilization}
\label{sec:post_sft}
Structured pruning modifies the effective network depth and can destabilize
both reasoning behavior and multimodal grounding.
To restore robustness, each pruned model is subjected to a brief supervised
fine-tuning (SFT) stage using a fixed mixture of mathematical and
general instruction data.
Applying an identical SFT protocol across all variants allows us to
attribute performance differences to pruning rather than downstream
adaptation.

Let the SFT dataset be
$\mathcal{D}_{\text{sft}} = \{(I, x, y)\}$,
where $I$ denotes the input image, $x$ the textual instruction,
and $y$ the target response.
We optimize the standard maximum-likelihood objective
\begin{equation}
\mathcal{L}_{\text{SFT}}(\theta)
= - \mathbb{E}_{(I,x,y) \sim \mathcal{D}_{\text{sft}}}
\left[ \log p_{\theta}(y \mid I, x) \right],
\label{eq:sft}
\end{equation}
where $p_{\theta}$ is the conditional output distribution of the pruned
model.
This objective encourages accurate instruction following conditioned on
both visual and textual inputs, and empirically mitigates performance
degradation introduced by structural pruning.

\section{Experiments and Dataset}

\subsection{Dataset}
\label{sec:datasets}

Our experiments draw on a mixture of proprietary and public multimodal datasets that collectively cover real-world homework photos, general vision-language tasks, scientific diagrams, chart reasoning, and fine-grained visual perception. Below we summarize each dataset, its size, and its role in training or evaluation.

\paragraph{Snapask.}
Snapask is a proprietary collection of roughly 300{,}000 real-world homework photographs spanning algebra, geometry, and general mathematics. Each image includes a curated GPT\mbox{-}4o \citep{openai_hello_gpt4o_2024} chain-of-thought solution. Following our internal data protocol, we use 100{,}000 examples for training and reserve a held-out evaluation set of 2{,}540 questions. Snapask provides the primary source of mathematical supervision for instruction tuning.

\paragraph{LLaVA-OneVision.}
LLaVA-OneVision contains 105{,}000 multimodal instruction-following samples covering charts, scientific figures, compositional reasoning, scene description, and VQA. We adopt the official 100k training split and the 5k held-out test split. This dataset supplies broad general-purpose vision-language coverage complementary to Snapask \citep{xie2025region,li2024llavaonevision,LLaVA-OneVision-1.5}.

\paragraph{NuminaMath-CoT.}
NuminaMath-CoT \citep{numina_math_datasets} consists of high-difficulty mathematical problems, including Olympiad-style and competition-level items paired with chain-of-thought solutions. To assess robustness under domain shift, we evaluate on a curated set of 500 items. Following \citep{khaki2026vistira}, all problems originally provided in text form are rendered as images.

\paragraph{ChartQA.}
ChartQA \citep{masry2022chartqa} evaluates chart and data visualization reasoning. It contains 9.6k human-authored questions and 23.1k automatically generated questions spanning bar charts, line plots, and statistical graphics. We use the standard 2{,}500-sample test split.

\paragraph{ScienceQA.}
ScienceQA \citep{lu2022scienceqa} includes 21{,}208 multimodal science questions spanning biology, physics, earth science, and general STEM topics. Many questions contain diagrams or textbook-style illustrations. We use the official 2{,}017-example evaluation split.

\paragraph{RealWorldQA.}
RealWorldQA \citep{realworldqa_xai} provides 765 real-world images paired with spatial or geometric reasoning questions. We evaluate on the full test set to measure robustness in real-world spatial understanding.

\paragraph{VStar.}
VStar \citep{wu2024vstar} is a fine-grained visual perception benchmark containing 191 tasks designed to evaluate subtle spatial relations, small-target identification, and fine-detail recognition in cluttered, high-resolution scenes.

\begin{table}[t]
\centering
\small
\setlength{\tabcolsep}{7pt}
\renewcommand{\arraystretch}{1.12}
\begin{tabular}{lcp{6.3cm}}
\hline
\textbf{Dataset} & \textbf{Size} & \textbf{Purpose} \\
\hline
Snapask & 300K total (100K train; 2{,}540 eval) &
Real-world homework photos with CoT solutions; primary math supervision. \\

LLaVA-OneVision & 105K total (100K train; 5K eval) &
Broad multimodal instruction tuning: charts, diagrams, scenes, VQA. \\

NuminaMath-CoT & 500 eval samples &
High-difficulty math reasoning; domain-shift robustness. \\

ChartQA & 32.7K questions (2{,}500 eval) &
Reasoning over charts and structured data visualizations. \\

ScienceQA & 21{,}208 total (2{,}017 eval) &
Multimodal science questions with diagrams and multi-step reasoning. \\

RealWorldQA & 765 images (full eval) &
Real-world spatial and geometric reasoning. \\

VStar & 191 tasks &
Fine-grained visual perception and small-target identification. \\
\hline
\end{tabular}
\caption{Training and evaluation datasets used in our experiments. Activation-analysis data is documented separately in Appendix~\ref{sec:activation_details}.}
\label{tab:dataset_summary}
\end{table}

\subsection{Training and Implementation Details}
\label{sec:train_setup}

All experiments are conducted using the Qwen3-VL-2B-Instruct and Qwen3-VL-4B-Instruct models. Structured pruning is applied exclusively to decoder layers, while the first and last decoder layers remain fixed. We evaluate three pruning budgets, 10\%, 25\%, and 40\%, measured as the proportion of pruneable decoder layers removed. To ensure a fair comparison, both baseline and pruned models share identical LoRA \citep{hu2022lora} configurations as well as identical optimization hyperparameters.

Training and evaluation are performed on 8 NVIDIA V100 GPUs (32\,GB each). The training mixture consists of 100k sampled Snapask examples combined with 100k instances from the LLaVA-OneVision training split, following the preprocessing pipeline described in Section~\ref{sec:datasets}. For held-out evaluation, we use the 5{,}000-sample LLaVA-OneVision test set, the Snapask validation subset (2{,}540 samples), and the competition-level NuminaMath-CoT benchmark.

We perform supervised fine-tuning (SFT) on the constructed training set using DeepSpeed ZeRO-3 \citep{ren2021deepspeed,rajbhandari2020zero,rasley2020deepspeed} to improve memory efficiency and scalability. Optimization follows a cosine learning-rate schedule with an initial learning rate of $2\times10^{-5}$, an effective batch size of 64, a single training epoch, weight decay of 0.1, and a maximum sequence length of 8{,}192 tokens. Parameter-efficient adaptation is implemented via LoRA \citep{hu2022lora} with rank $r=32$, $\alpha=64$, and dropout rate 0.05.

For broader multimodal evaluation, we additionally report results using the \texttt{lmms-eval} framework \citep{zhang2024lmms_eval} on the following benchmarks:
\begin{itemize}
    \item ChartQA \citep{masry2022chartqa}
    \item RealWorldQA \citep{realworldqa_xai}
    \item ScienceQA \citep{lu2022scienceqa}
    \item VStar \citep{wu2024vstar}
\end{itemize}

To compute accuracy on LLaVA-OneVision, Snapask, and NuminaMath-CoT, we use GPT-5 \citep{openai_gpt5_system_card_2025} to compare the model's predicted responses against the ground-truth answers.

\subsection{Baseline Pruning Methods}
\label{sec:baselines}

To contextualize the effectiveness of our domain-aware pruning strategy, we compare against three widely used baselines representing complementary pruning philosophies: representation-similarity pruning, structure-aware pruning, and uninformed random removal.

\paragraph{CKA-based pruning.}
We employ \emph{linear Centered Kernel Alignment (CKA)}~\cite{pmlr-v97-kornblith19a} to quantify 
how similarly two consecutive decoder layers represent their inputs. 
CKA is a representation-similarity metric that is invariant to isotropic scaling and 
orthogonal transformations, making it robust to the natural feature rotations that occur in 
deep networks~\cite{pmlr-v97-kornblith19a}.  In our setting, for each layer $\ell$, we construct a 
feature matrix $X_\ell \in \mathbb{R}^{9 \times d}$ by stacking the mean-pooled activation 
vectors from the nine probing tasks (five math tasks and four non-math tasks), treating each 
task as a pseudo-sample.  
After centering the features, we compute linear CKA between layers $\ell$ 
and $\ell{+}1$ as
\[
\mathrm{CKA}(X_\ell, X_{\ell+1}) 
= 
\frac{\left\|\tilde{X}_\ell^\top \tilde{X}_{\ell+1}\right\|_{F}^{2}}
     {\left\|\tilde{X}_\ell^\top \tilde{X}_\ell\right\|_{F}
      \left\|\tilde{X}_{\ell+1}^\top \tilde{X}_{\ell+1}\right\|_{F}}.
\]
A high CKA value indicates that the two layers encode highly aligned information and thus 
perform similar transformations on the probing tasks~\cite{pmlr-v97-kornblith19a}.  
We therefore interpret high CKA$(\ell,\ell{+}1)$ as evidence that layer $\ell{+}1$ is 
\emph{redundant} given layer $\ell$. Layers are ranked by redundancy and pruned from highest 
to lowest under the same budget and protection rules described in 
Section~\ref{sec:ranking}.

\paragraph{Interlace pruning.}
Interlace~\cite{Madinei2025INTERLACE} is a structure-aware depth pruning strategy designed 
to avoid the instability caused by removing many adjacent layers, a common failure mode of 
similarity-only methods. Interlace analyzes \emph{local triplets} of layers 
$(\ell,\ell{+}1,\ell{+}2)$, using the same nine pseudo-sample activations as in our CKA 
baseline. For each layer, we first compute a cosine similarity
\[
S_\ell = \frac{1}{9}\sum_{i=1}^{9} 
\cos\!\left(h_\ell^{(i)},\, h_{\ell+1}^{(i)}\right),
\]
where $h_\ell^{(i)}$ is the mean-pooled activation for task $i$.  
Each triplet receives a \emph{triplet redundancy score}
\[
R_{(\ell,\ell+1,\ell+2)} = \frac{S_\ell + S_{\ell+1}}{2},
\]
which captures how weakly the layers in that local region transform their 
representations~\cite{Madinei2025INTERLACE}.  
Triplets are then sorted by $R$ and selected greedily under a strict 
\emph{non-overlap constraint}, ensuring that no two chosen triplets touch.  
Within each selected triplet, Interlace removes exactly one layer---the more redundant of 
$\ell$ or $\ell{+}1$---while always preserving $\ell{+}2$ as a stable \emph{anchor layer}.  
This explicit spacing constraint prevents the formation of long contiguous holes in the 
network, making Interlace particularly effective at higher pruning budgets where structural 
collapse otherwise becomes a limiting factor~\cite{Madinei2025INTERLACE}.

\paragraph{Random pruning.}
Random pruning selects $K$ pruneable layers uniformly at random (seeded for reproducibility). Despite its simplicity, prior work has shown that random removal can be surprisingly competitive in large models, making it a necessary sanity-check baseline.

All baselines are followed by the same post-pruning SFT procedure described in Section~\ref{sec:post_sft}, ensuring that differences in performance stem solely from the pruning strategy rather than subsequent adaptation.

\section{Results}
\label{sec:results}

We report accuracy across math benchmarks (Snapask, NuminaMath) and general vision-language benchmarks (LLaVA-OneVision, ChartQA, RealWorldQA, ScienceQA, VStar). All deltas are computed relative to the \emph{unpruned baseline model without fine-tuning}. Appendix Table~\ref{tab:baseline_sft_reference} reports the corresponding baseline+SFT values.

\paragraph{Cell format.}
Each table cell is reported as
\[
\text{Acc} \; (\Delta_{\text{abs}}, \Delta_{\text{rel}}),
\qquad
\Delta_{\text{abs}}=\text{Acc}-\text{Acc}_{\text{Baseline}},
\qquad
\Delta_{\text{rel}}=100\cdot \frac{\Delta_{\text{abs}}}{\text{Acc}_{\text{Baseline}}}.
\]
Bold indicates the highest accuracy among pruned variants at that budget.

\begin{figure*}[t]
\centering
\includegraphics[scale=0.7]{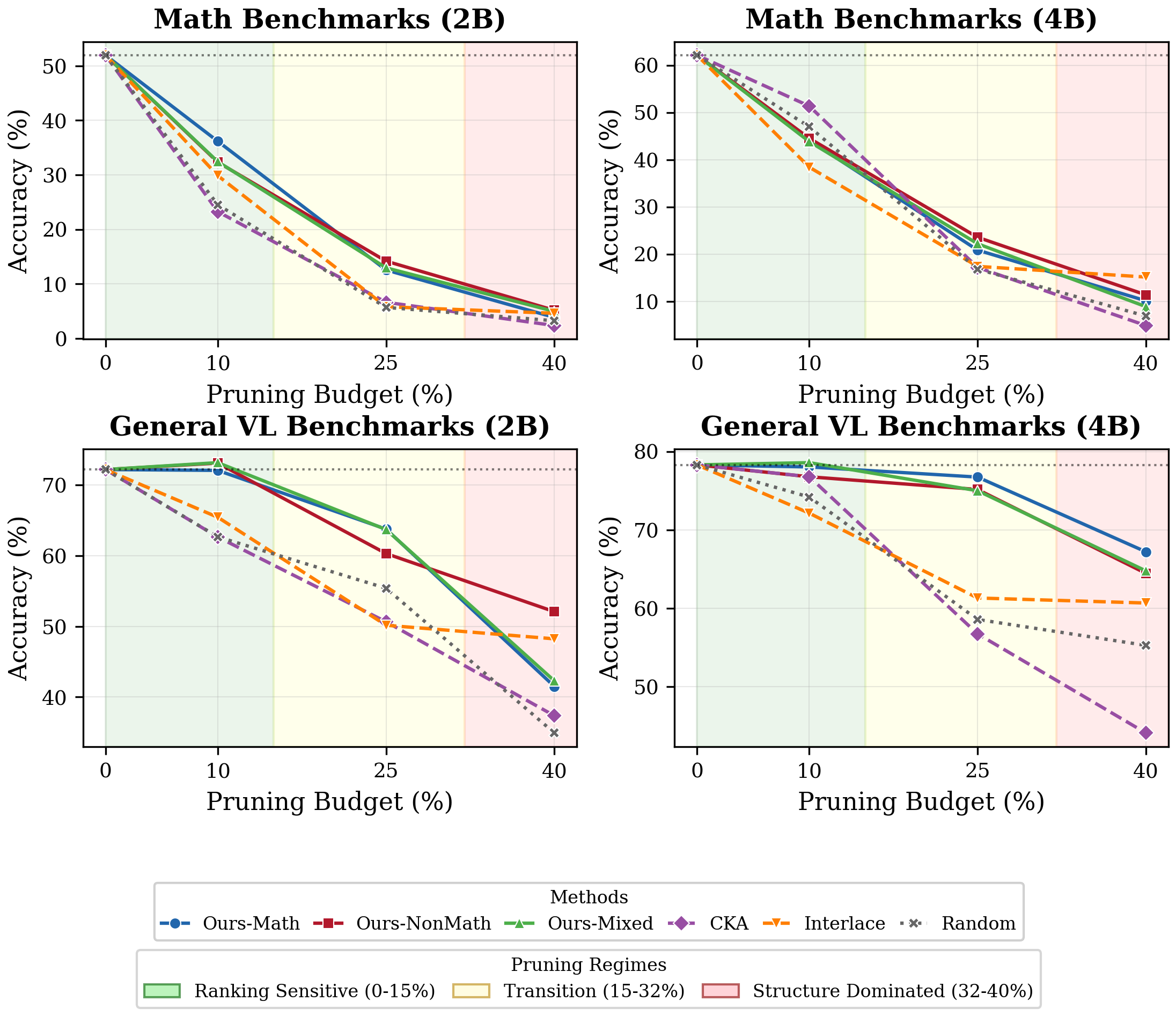}
\caption{Accuracy versus pruning budget across domains and model sizes. Shaded regions highlight the ranking-sensitive regime (0--15\%), transition regime (15--32\%), and structure-dominated regime (32--40\%). Domain-aware methods separate most strongly at low budgets, while structural continuity becomes limiting at high budgets.}
\label{fig:regime_transitions}
\end{figure*}

\paragraph{High-level trends.}
Figure~\ref{fig:regime_transitions} visualizes the three pruning regimes underlying our results. At low budgets, which layers are removed matters most; at high budgets the model becomes structure-limited. These regimes are visible across both model scales and both domains.

\subsection{Math domain: Snapask and NuminaMath}
\label{sec:results_math}

Snapask consists of real-world homework photos, where problems are often accompanied by diagrams, tables, and other visual elements. NuminaMath contains competition-style math problems that exhibit significant domain shift. Across both benchmarks, we observe a clear regime structure: domain-aware ranking is most effective under low-budget settings, whereas structural constraints become dominant at higher budgets.

\begin{table*}[t]
\centering
\small
\setlength{\tabcolsep}{3.5pt}
\renewcommand{\arraystretch}{1.15}
\caption{Math results for Qwen3-VL-2B. Deltas are relative to the unpruned baseline without fine-tuning.}
\label{tab:math_2b}

\vspace{1mm}
\textbf{Budget 10\% (ranking-sensitive regime)}\\
\resizebox{\textwidth}{!}{%
\begin{tabular}{lcccccc}
\hline
\textbf{Dataset} & \textbf{Ours-Math} & \textbf{Ours-NonMath} & \textbf{Ours-Mixed} & \textbf{CKA} & \textbf{Interlace} & \textbf{Random} \\
\hline
Snapask & \textbf{28.15 (-12.32, -30.4\%)} & 24.69 (-15.78, -39.0\%) & 24.88 (-15.59, -38.5\%) & 15.98 (-24.49, -60.5\%) & 20.75 (-19.72, -48.7\%) & 17.09 (-23.38, -57.8\%) \\
NuminaMath & \textbf{44.18 (-19.27, -30.4\%)} & 39.96 (-23.49, -37.0\%) & 39.96 (-23.49, -37.0\%) & 30.52 (-32.93, -51.9\%) & 38.96 (-24.49, -38.6\%) & 31.93 (-31.52, -49.7\%) \\
\hline
\end{tabular}%
}

\vspace{2mm}
\textbf{Budget 25\% (transition regime)}\\
\resizebox{\textwidth}{!}{%
\begin{tabular}{lcccccc}
\hline
\textbf{Dataset} & \textbf{Ours-Math} & \textbf{Ours-NonMath} & \textbf{Ours-Mixed} & \textbf{CKA} & \textbf{Interlace} & \textbf{Random} \\
\hline
Snapask & 9.21 (-31.26, -77.2\%) & \textbf{11.06 (-29.41, -72.7\%)} & 9.61 (-30.86, -76.3\%) & 4.80 (-35.67, -88.1\%) & 5.17 (-35.30, -87.2\%) & 4.72 (-35.75, -88.3\%) \\
NuminaMath & 15.86 (-47.59, -75.0\%) & \textbf{17.27 (-46.18, -72.8\%)} & 16.27 (-47.18, -74.4\%) & 8.43 (-55.02, -86.7\%) & 6.43 (-57.02, -89.9\%) & 6.63 (-56.82, -89.6\%) \\
\hline
\end{tabular}%
}

\vspace{2mm}
\textbf{Budget 40\% (structure-dominated regime)}\\
\resizebox{\textwidth}{!}{%
\begin{tabular}{lcccccc}
\hline
\textbf{Dataset} & \textbf{Ours-Math} & \textbf{Ours-NonMath} & \textbf{Ours-Mixed} & \textbf{CKA} & \textbf{Interlace} & \textbf{Random} \\
\hline
Snapask & 3.46 (-37.01, -91.5\%) & 4.06 (-36.41, -90.0\%) & \textbf{4.65 (-35.82, -88.5\%)} & 2.48 (-37.99, -93.9\%) & 4.57 (-35.90, -88.7\%) & 2.44 (-38.03, -94.0\%) \\
NuminaMath & 4.22 (-59.23, -93.3\%) & \textbf{6.22 (-57.23, -90.2\%)} & 5.22 (-58.23, -91.8\%) & 2.21 (-61.24, -96.5\%) & 4.62 (-58.83, -92.7\%) & 4.02 (-59.43, -93.7\%) \\
\hline
\end{tabular}%
}
\end{table*}

\begin{table*}[t]
\centering
\small
\setlength{\tabcolsep}{3.5pt}
\renewcommand{\arraystretch}{1.15}
\caption{Math results for Qwen3-VL-4B. Deltas are relative to the unpruned baseline without fine-tuning.}
\label{tab:math_4b}

\vspace{1mm}
\textbf{Budget 10\% (ranking-sensitive regime)}\\
\resizebox{\textwidth}{!}{%
\begin{tabular}{lcccccc}
\hline
\textbf{Dataset} & \textbf{Ours-Math} & \textbf{Ours-NonMath} & \textbf{Ours-Mixed} & \textbf{CKA} & \textbf{Interlace} & \textbf{Random} \\
\hline
Snapask & 35.83 (-14.92, -29.4\%) & 33.54 (-17.21, -33.9\%) & 33.23 (-17.52, -34.5\%) & \textbf{40.24 (-10.51, -20.7\%)} & 28.82 (-21.93, -43.2\%) & 37.09 (-13.66, -26.9\%) \\
NuminaMath & 52.41 (-21.08, -28.7\%) & 55.50 (-17.99, -24.5\%) & 54.42 (-19.07, -25.9\%) & \textbf{62.45 (-11.04, -15.0\%)} & 47.99 (-25.50, -34.7\%) & 56.83 (-16.66, -22.7\%) \\
\hline
\end{tabular}%
}

\vspace{2mm}
\textbf{Budget 25\% (transition regime)}\\
\resizebox{\textwidth}{!}{%
\begin{tabular}{lcccccc}
\hline
\textbf{Dataset} & \textbf{Ours-Math} & \textbf{Ours-NonMath} & \textbf{Ours-Mixed} & \textbf{CKA} & \textbf{Interlace} & \textbf{Random} \\
\hline
Snapask & 15.98 (-34.77, -68.5\%) & \textbf{18.19 (-32.56, -64.2\%)} & 16.26 (-34.49, -68.0\%) & 12.44 (-38.31, -75.5\%) & 12.09 (-38.66, -76.2\%) & 11.30 (-39.45, -77.7\%) \\
NuminaMath & 25.70 (-47.79, -65.0\%) & \textbf{28.92 (-44.57, -60.6\%)} & 28.11 (-45.38, -61.7\%) & 21.89 (-51.60, -70.2\%) & 22.69 (-50.80, -69.1\%) & 22.29 (-51.20, -69.7\%) \\
\hline
\end{tabular}%
}

\vspace{2mm}
\textbf{Budget 40\% (structure-dominated regime)}\\
\resizebox{\textwidth}{!}{%
\begin{tabular}{lcccccc}
\hline
\textbf{Dataset} & \textbf{Ours-Math} & \textbf{Ours-NonMath} & \textbf{Ours-Mixed} & \textbf{CKA} & \textbf{Interlace} & \textbf{Random} \\
\hline
Snapask & 9.76 (-40.99, -80.8\%) & 9.06 (-41.69, -82.1\%) & 7.68 (-43.07, -84.9\%) & 3.70 (-47.05, -92.7\%) & \textbf{10.79 (-39.96, -78.7\%)} & 5.51 (-45.24, -89.1\%) \\
NuminaMath & 10.44 (-63.05, -85.8\%) & 13.65 (-59.84, -81.4\%) & 10.04 (-63.45, -86.3\%) & 6.02 (-67.47, -91.8\%) & \textbf{19.48 (-54.01, -73.5\%)} & 8.23 (-65.26, -88.8\%) \\
\hline
\end{tabular}%
}
\end{table*}

\paragraph{Math insights across regimes.}
For the 2B model at 10\%, domain-aware ranking is consistently the most stable. At higher budgets, all methods deteriorate sharply, reflecting the depth sensitivity of math reasoning. For the 4B model, 10\% remains competitive, but at 40\%, Interlace becomes the most stable due to its spaced layer deletions, which avoid contiguous holes that disrupt long-range residual flow.

Appendix Figures~\ref{fig:relative_drop_bars}, \ref{fig:per_benchmark_curves_2b}, and \ref{fig:per_benchmark_curves_4b} provide a compact view of relative degradation and per-benchmark curves.

\begin{figure*}[t]
    \centering
    \includegraphics[scale=0.5]{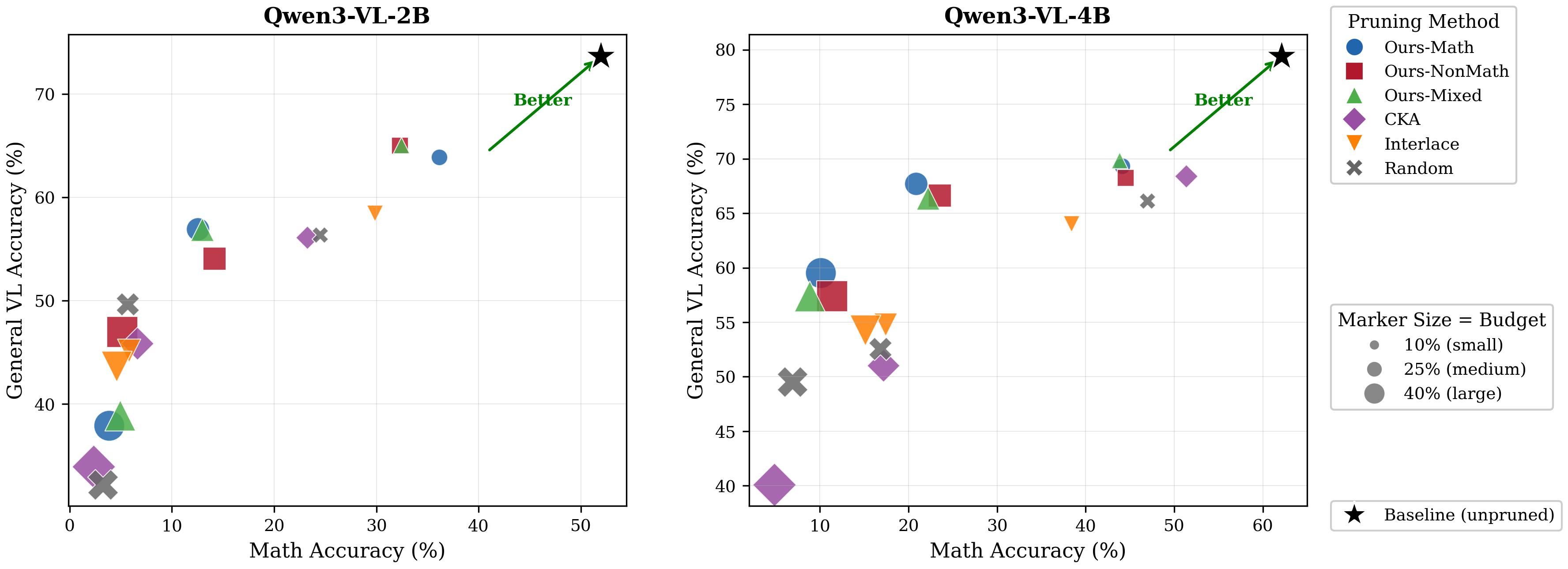}
    \caption{
    Trade-off between Math accuracy (x-axis) and General Vision-Language accuracy 
    (y-axis) for Qwen3-VL-2B (left) and Qwen3-VL-4B (right). Each point represents 
    a pruning method at a given budget; marker size increases with pruning budget 
    (10\%, 25\%, 40\%). The black star denotes the unpruned baseline. Domain-aware 
    methods (circles) lie on or near the Pareto frontier, achieving the best 
    balance between math and general VL performance. CKA and Random pruning are 
    consistently dominated, while Interlace maintains general VL stability but 
    exhibits steeper math degradation at higher budgets.
    }
    \label{fig:math_general_tradeoff}
\end{figure*}
\paragraph{Math--General VL trade-off.}
To summarize how pruning affects mathematical reasoning and general multimodal 
ability jointly, Figure~\ref{fig:math_general_tradeoff} plots Math accuracy 
against General VL accuracy for all methods and pruning budgets. Each point 
corresponds to a (method, budget) pair, with marker size indicating pruning 
budget. The unpruned baseline appears in the upper-right corner. Domain-aware 
methods trace a Pareto-efficient frontier: Ours-Math shifts rightward 
(stronger math retention), Ours-NonMath shifts upward (stronger general VL 
retention), and Ours-Mixed consistently lies near the diagonal frontier, 
offering the best balance. In contrast, CKA and Random pruning are dominated 
across both axes, while Interlace maintains reasonable general VL performance 
but suffers sharper math degradation at higher budgets. This trade-off view 
highlights that domain-aware ranking provides the most favorable balance when 
both math and general multimodal capability must be preserved.

\subsection{General vision-language: LLaVA-OneVision, ChartQA, RealWorldQA, ScienceQA, VStar}
\label{sec:results_general}

We evaluate general vision-language ability using LLaVA-OneVision and four \texttt{lmms-eval} benchmarks \citep{zhang2024lmms_eval}, each targeting different skills: ChartQA (structured chart parsing), RealWorldQA (spatial grounding), ScienceQA (diagram reasoning), and VStar (fine-grained perception). The transition from ranking sensitivity to structure sensitivity is visible across all tasks as the pruning budget increases.

\begin{table*}[t]
\centering
\scriptsize
\setlength{\tabcolsep}{3pt}
\renewcommand{\arraystretch}{1.12}
\caption{General vision-language results for Qwen3-VL-2B. Deltas are relative to the unpruned baseline without fine-tuning. Scores for ChartQA, RealWorldQA, ScienceQA, and VStar are obtained using \texttt{lmms-eval} \citep{zhang2024lmms_eval}.}
\label{tab:general_2b}

\vspace{1mm}
\textbf{Budget 10\% (ranking-sensitive regime)}\\
\resizebox{\textwidth}{!}{%
\begin{tabular}{lcccccc}
\hline
\textbf{Dataset} & \textbf{Ours-Math} & \textbf{Ours-NonMath} & \textbf{Ours-Mixed} & \textbf{CKA} & \textbf{Interlace} & \textbf{Random} \\
\hline
LLaVA-OneVision & 60.74 (+1.62, +2.7\%) & \textbf{60.76 (+1.64, +2.8\%)} & \textbf{60.76 (+1.64, +2.8\%)} & 53.84 (-5.28, -8.9\%) & 57.70 (-1.42, -2.4\%) & 54.38 (-4.74, -8.0\%) \\
ChartQA & 31.28 (-48.44, -60.8\%) & \textbf{32.68 (-47.04, -59.0\%)} & 32.60 (-47.12, -59.1\%) & 30.08 (-49.64, -62.3\%) & 30.60 (-49.12, -61.6\%) & 31.16 (-48.56, -60.9\%) \\
RealWorldQA & 65.10 (-0.13, -0.2\%) & \textbf{67.45 (+2.22, +3.4\%)} & 67.32 (+2.09, +3.2\%) & 59.87 (-5.36, -8.2\%) & 61.70 (-3.53, -5.4\%) & 61.44 (-3.79, -5.8\%) \\
ScienceQA & \textbf{88.15 (+1.68, +1.9\%)} & 87.31 (+0.84, +1.0\%) & 87.70 (+1.23, +1.4\%) & 74.07 (-12.40, -14.3\%) & 73.82 (-12.65, -14.6\%) & 70.50 (-15.97, -18.5\%) \\
VStar & 74.35 (-3.66, -4.7\%) & \textbf{76.96 (-1.05, -1.3\%)} & \textbf{76.96 (-1.05, -1.3\%)} & 62.83 (-15.18, -19.5\%) & 68.59 (-9.42, -12.1\%) & 64.40 (-13.61, -17.4\%) \\
\hline
\end{tabular}%
}

\vspace{2mm}
\textbf{Budget 25\% (transition regime)}\\
\resizebox{\textwidth}{!}{%
\begin{tabular}{lcccccc}
\hline
\textbf{Dataset} & \textbf{Ours-Math} & \textbf{Ours-NonMath} & \textbf{Ours-Mixed} & \textbf{CKA} & \textbf{Interlace} & \textbf{Random} \\
\hline
LLaVA-OneVision & \textbf{50.62 (-8.50, -14.4\%)} & 50.52 (-8.60, -14.5\%) & 50.56 (-8.56, -14.5\%) & 43.70 (-15.42, -26.1\%) & 42.47 (-16.65, -28.2\%) & 46.80 (-12.32, -20.8\%) \\
ChartQA & 29.56 (-50.16, -62.9\%) & 29.04 (-50.68, -63.6\%) & \textbf{29.64 (-50.08, -62.8\%)} & 26.68 (-53.04, -66.5\%) & 25.24 (-54.48, -68.3\%) & 26.80 (-52.92, -66.4\%) \\
RealWorldQA & \textbf{58.95 (-6.28, -9.6\%)} & 58.43 (-6.80, -10.4\%) & \textbf{58.95 (-6.28, -9.6\%)} & 50.20 (-15.03, -23.0\%) & 47.84 (-17.39, -26.7\%) & 55.29 (-9.94, -15.2\%) \\
ScienceQA & \textbf{75.90 (-10.57, -12.2\%)} & 68.42 (-18.05, -20.9\%) & 75.71 (-10.76, -12.4\%) & 58.95 (-27.52, -31.8\%) & 59.54 (-26.93, -31.1\%) & 61.38 (-25.09, -29.0\%) \\
VStar & \textbf{69.63 (-8.38, -10.7\%)} & 63.87 (-14.14, -18.1\%) & \textbf{69.63 (-8.38, -10.7\%)} & 49.74 (-28.27, -36.2\%) & 50.79 (-27.22, -34.9\%) & 58.12 (-19.89, -25.5\%) \\
\hline
\end{tabular}%
}

\vspace{2mm}
\textbf{Budget 40\% (structure-dominated regime)}\\
\resizebox{\textwidth}{!}{%
\begin{tabular}{lcccccc}
\hline
\textbf{Dataset} & \textbf{Ours-Math} & \textbf{Ours-NonMath} & \textbf{Ours-Mixed} & \textbf{CKA} & \textbf{Interlace} & \textbf{Random} \\
\hline
LLaVA-OneVision & 37.32 (-21.80, -36.9\%) & \textbf{43.22 (-15.90, -26.9\%)} & 40.38 (-18.74, -31.7\%) & 34.58 (-24.54, -41.5\%) & 40.94 (-18.18, -30.8\%) & 32.64 (-26.48, -44.8\%) \\
ChartQA & 23.72 (-56.00, -70.2\%) & \textbf{26.56 (-53.16, -66.7\%)} & 25.52 (-54.20, -68.0\%) & 20.40 (-59.32, -74.4\%) & 25.36 (-54.36, -68.2\%) & 21.48 (-58.24, -73.1\%) \\
RealWorldQA & 45.88 (-19.35, -29.7\%) & \textbf{52.81 (-12.42, -19.0\%)} & 47.19 (-18.04, -27.7\%) & 40.26 (-24.97, -38.3\%) & 48.24 (-16.99, -26.0\%) & 36.73 (-28.50, -43.7\%) \\
ScienceQA & 42.94 (-43.53, -50.3\%) & \textbf{57.31 (-29.16, -33.7\%)} & 44.37 (-42.10, -48.7\%) & 35.25 (-51.22, -59.2\%) & 57.16 (-29.31, -33.9\%) & 35.15 (-51.32, -59.4\%) \\
VStar & 39.79 (-38.22, -49.0\%) & \textbf{54.97 (-23.04, -29.5\%)} & 37.17 (-40.84, -52.4\%) & 39.27 (-38.74, -49.7\%) & 46.60 (-31.41, -40.3\%) & 35.08 (-42.93, -55.0\%) \\
\hline
\end{tabular}%
}
\end{table*}

\begin{table*}[t]
\centering
\scriptsize
\setlength{\tabcolsep}{3pt}
\renewcommand{\arraystretch}{1.12}
\caption{General vision-language results for Qwen3-VL-4B. Deltas are relative to the unpruned baseline without fine-tuning. Scores for ChartQA, RealWorldQA, ScienceQA, and VStar are obtained using \texttt{lmms-eval} \citep{zhang2024lmms_eval}.}
\label{tab:general_4b}

\vspace{1mm}
\textbf{Budget 10\% (ranking-sensitive regime)}\\
\resizebox{\textwidth}{!}{%
\begin{tabular}{lcccccc}
\hline
\textbf{Dataset} & \textbf{Ours-Math} & \textbf{Ours-NonMath} & \textbf{Ours-Mixed} & \textbf{CKA} & \textbf{Interlace} & \textbf{Random} \\
\hline
LLaVA-OneVision & 66.16 (-1.10, -1.6\%) & 64.62 (-2.64, -3.9\%) & 66.02 (-1.24, -1.8\%) & \textbf{66.82 (-0.44, -0.7\%)} & 63.14 (-4.12, -6.1\%) & 64.56 (-2.70, -4.0\%) \\
ChartQA & 34.48 (-49.52, -59.0\%) & 34.12 (-49.88, -59.4\%) & \textbf{35.04 (-48.96, -58.3\%)} & 34.84 (-49.16, -58.5\%) & 31.56 (-52.44, -62.4\%) & 33.84 (-50.16, -59.7\%) \\
RealWorldQA & 71.63 (+0.13, +0.2\%) & 69.54 (-1.96, -2.7\%) & \textbf{72.29 (+0.79, +1.1\%)} & 71.11 (-0.39, -0.5\%) & 64.97 (-6.53, -9.1\%) & 67.97 (-3.53, -4.9\%) \\
ScienceQA & 93.16 (+0.45, +0.5\%) & 91.77 (-0.94, -1.0\%) & \textbf{93.26 (+0.55, +0.6\%)} & 86.47 (-6.24, -6.7\%) & 86.61 (-6.10, -6.6\%) & 86.22 (-6.49, -7.0\%) \\
VStar & 81.15 (-0.53, -0.6\%) & 81.15 (-0.53, -0.6\%) & \textbf{82.72 (+1.04, +1.3\%)} & \textbf{82.72 (+1.04, +1.3\%)} & 73.82 (-7.86, -9.6\%) & 78.01 (-3.67, -4.5\%) \\
\hline
\end{tabular}%
}

\vspace{2mm}
\textbf{Budget 25\% (transition regime)}\\
\resizebox{\textwidth}{!}{%
\begin{tabular}{lcccccc}
\hline
\textbf{Dataset} & \textbf{Ours-Math} & \textbf{Ours-NonMath} & \textbf{Ours-Mixed} & \textbf{CKA} & \textbf{Interlace} & \textbf{Random} \\
\hline
LLaVA-OneVision & \textbf{61.78 (-5.48, -8.1\%)} & 61.32 (-5.94, -8.8\%) & 60.78 (-6.48, -9.6\%) & 51.86 (-15.40, -22.9\%) & 55.20 (-12.06, -17.9\%) & 51.76 (-15.50, -23.0\%) \\
ChartQA & 31.60 (-52.40, -62.4\%) & \textbf{32.40 (-51.60, -61.4\%)} & 32.00 (-52.00, -61.9\%) & 28.24 (-55.76, -66.4\%) & 28.64 (-55.36, -65.9\%) & 28.60 (-55.40, -66.0\%) \\
RealWorldQA & \textbf{70.46 (-1.04, -1.5\%)} & 68.76 (-2.74, -3.8\%) & 68.50 (-3.00, -4.2\%) & 57.91 (-13.59, -19.0\%) & 62.88 (-8.62, -12.1\%) & 58.43 (-13.07, -18.3\%) \\
ScienceQA & \textbf{91.97 (-0.74, -0.8\%)} & 90.43 (-2.28, -2.5\%) & 91.62 (-1.09, -1.2\%) & 61.68 (-31.03, -33.5\%) & 68.02 (-24.69, -26.6\%) & 66.53 (-26.18, -28.2\%) \\
VStar & \textbf{82.72 (+1.04, +1.3\%)} & 80.10 (-1.58, -1.9\%) & 79.06 (-2.62, -3.2\%) & 55.50 (-26.18, -32.1\%) & 59.16 (-22.52, -27.6\%) & 57.59 (-24.09, -29.5\%) \\
\hline
\end{tabular}%
}

\vspace{2mm}
\textbf{Budget 40\% (structure-dominated regime)}\\
\resizebox{\textwidth}{!}{%
\begin{tabular}{lcccccc}
\hline
\textbf{Dataset} & \textbf{Ours-Math} & \textbf{Ours-NonMath} & \textbf{Ours-Mixed} & \textbf{CKA} & \textbf{Interlace} & \textbf{Random} \\
\hline
LLaVA-OneVision & 53.88 (-13.38, -19.9\%) & 54.00 (-13.26, -19.7\%) & 50.72 (-16.54, -24.6\%) & 40.72 (-26.54, -39.5\%) & \textbf{54.26 (-13.00, -19.3\%)} & 46.64 (-20.62, -30.7\%) \\
ChartQA & 28.92 (-55.08, -65.6\%) & \textbf{29.04 (-54.96, -65.4\%)} & 27.96 (-56.04, -66.7\%) & 24.20 (-59.80, -71.2\%) & 28.64 (-55.36, -65.9\%) & 26.64 (-57.36, -68.3\%) \\
RealWorldQA & \textbf{63.01 (-8.49, -11.9\%)} & 61.70 (-9.80, -13.7\%) & 60.92 (-10.58, -14.8\%) & 46.27 (-25.23, -35.3\%) & 62.88 (-8.62, -12.1\%) & 56.73 (-14.77, -20.7\%) \\
ScienceQA & \textbf{81.76 (-10.95, -11.8\%)} & 73.53 (-19.18, -20.7\%) & 79.38 (-13.33, -14.4\%) & 50.67 (-42.04, -45.3\%) & 67.97 (-24.74, -26.7\%) & 64.85 (-27.86, -30.1\%) \\
VStar & \textbf{70.16 (-11.52, -14.1\%)} & 68.59 (-13.09, -16.0\%) & 68.06 (-13.62, -16.7\%) & 38.74 (-42.94, -52.6\%) & 57.59 (-24.09, -29.5\%) & 52.88 (-28.80, -35.3\%) \\
\hline
\end{tabular}%
}
\end{table*}

\paragraph{Task-level insights beyond averages.}

\vspace{1em}

\textbf{ChartQA.} ChartQA is consistently fragile under our fixed post-pruning SFT mixture, producing large absolute drops relative to the no-SFT baseline across all pruning strategies and budgets. Appendix Table~\ref{tab:baseline_sft_reference} shows that the unpruned baseline+SFT itself is much lower than the baseline on ChartQA, indicating that the primary driver is the distribution shift induced by the stabilization SFT rather than the specific pruning strategy. Under this constraint, mixed and non-math rankings are still the most stable among pruned models at each budget, suggesting that preserving general vision-language layers is preferable when chart-specific data is not present in the SFT mixture.

\textbf{RealWorldQA.} RealWorldQA remains close to baseline at 10\% for both scales, with mixed or non-math ranking performing best. At higher budgets, the most stable outcomes shift toward strategies that avoid removing early-to-mid layers, consistent with RealWorldQA requiring robust spatial grounding and cross-modal alignment. This aligns with the fact that our math-aware pruning tends to target later layers, while CKA removes earlier blocks more aggressively.

\textbf{ScienceQA.} ScienceQA behaves more like a reasoning benchmark than a pure perception benchmark: at 10\% the best methods often match or slightly exceed the baseline, and at 25\% the math-aware variant is consistently strongest for both 2B and 4B. This is consistent with ScienceQA requiring multi-step language reasoning conditioned on diagrams, benefiting from preserving reasoning-critical mid-depth computation while pruning redundant late layers. The layer logs confirm that math-aware pruning concentrates deletions in late layers for both model sizes at small budgets.

\textbf{VStar.} VStar is sensitive to aggressive pruning but can tolerate mild pruning. At 10\%, mixed and CKA reach the best scores in 4B, while non-math and mixed are most stable in 2B. This suggests that fine-grained perception benefits from preserving early and mid layers and that removing redundant late layers can be safe, but once pruning becomes aggressive the task becomes structure-limited.

Appendix Figures~\ref{fig:per_benchmark_curves_2b}--\ref{fig:per_benchmark_curves_4b} show full per-benchmark curves.

\subsection{Layer removal patterns and regime transitions}
\label{sec:results_layer_patterns}

\begin{figure*}[t]
    \centering
    \includegraphics[scale=0.4]{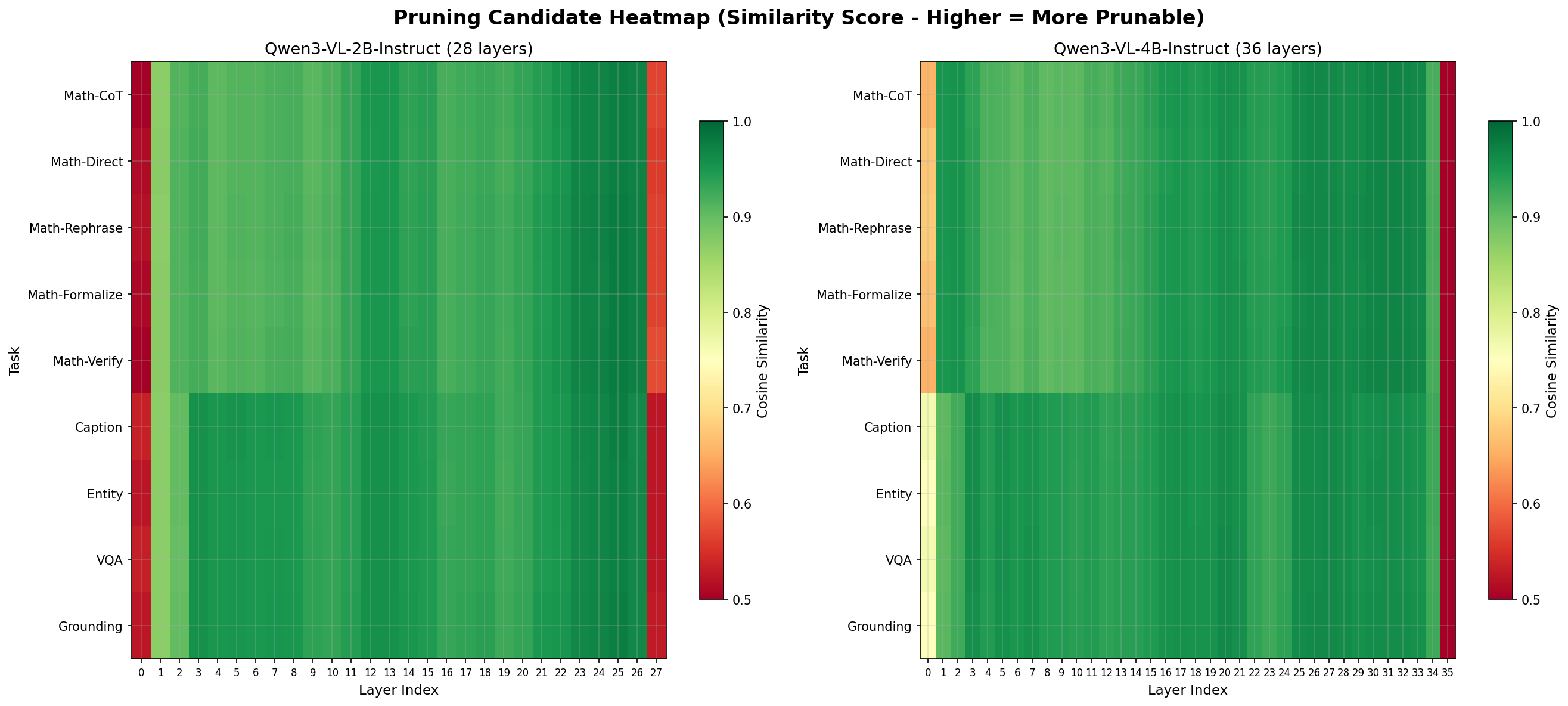}
    \caption{
    Pruning candidate heatmaps for Qwen3-VL-2B (left) and Qwen3-VL-4B (right). 
    Each cell shows the cosine similarity between a layer's input and output 
    activations for a given probing subtask. Higher similarity indicates weaker 
    transformation and thus higher pruneability. Math subtasks (top rows) and 
    non-math subtasks (bottom rows) exhibit distinct activation patterns, 
    motivating domain-aware layer ranking.
    }
    \label{fig:pruning_heatmap}
\end{figure*}

To understand why pruning strategies behave differently across budgets, we first examine 
how strongly each decoder layer transforms its representations under the probing tasks used 
to compute redundancy scores. Figure~\ref{fig:pruning_heatmap} visualizes the input--output 
cosine similarity for every layer across all math and non-math subtasks. Layers with higher 
similarity perform weaker updates and are therefore more pruneable under our 
criterion. The heatmaps reveal a consistent structure across both model sizes: math subtasks 
elicit stronger activations in mid--late layers, while non-math subtasks rely more on early--mid layers. 
These domain-specific activation patterns directly shape the behavior of our math-aware, 
non-math-aware, and mixed rankings.

\begin{figure*}[t]
\centering
\includegraphics[width=\textwidth]{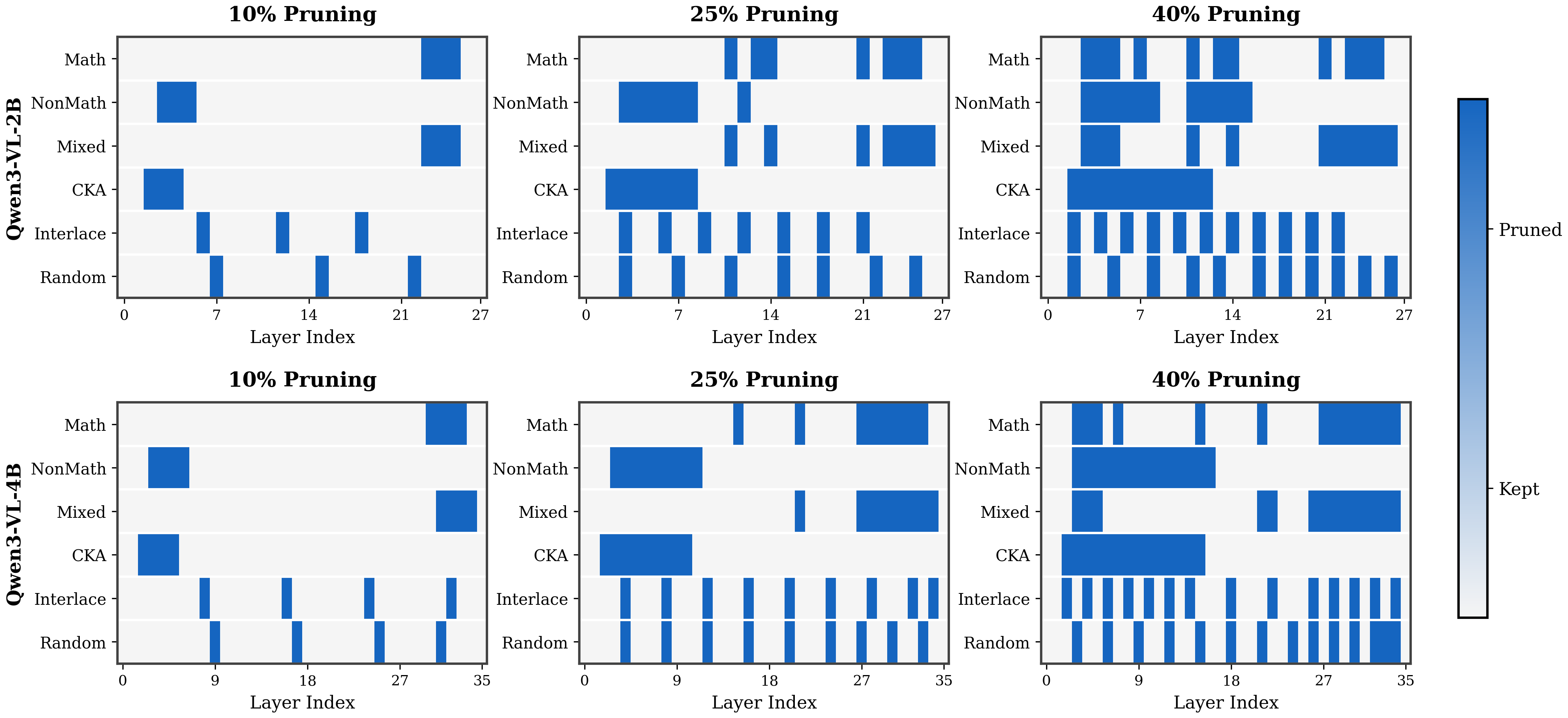}
\caption{Decoder layer removal patterns across methods, budgets, and model sizes. Dark cells indicate pruned layers; light cells indicate retained layers. Domain-aware methods produce structured patterns, while Interlace enforces spaced deletions and Random remains unstructured.}
\label{fig:layer_removal_heatmap}
\end{figure*}

Figure~\ref{fig:layer_removal_heatmap} shows the exact pruned layers for each method, budget, and model scale. The mid-depth deletions vary significantly across strategies.

Math-aware ranking concentrates deletions in the later layers at small budgets, while non-math-aware and mixed rankings begin removing earlier blocks as the budget grows. CKA consistently targets early layers, whereas Interlace enforces evenly spaced deletions across depth.

These patterns explain the regimes in Figure~\ref{fig:regime_transitions}. At low budgets, ranking is most effective because it avoids deleting critical computation. At high budgets, the model becomes structure-limited, and strategies that preserve continuity---especially Interlace---become more stable.

Random pruning produces the largest drops across domains because it deletes early or mid-depth blocks without structural safeguards, often producing irregular gaps.

\bibliographystyle{unsrt}  
\bibliography{references}  

\appendix
\section{Appendix}
\subsection{Activation Capture and Processing Details}
\label{sec:activation_details}

This section provides additional details on the activation extraction pipeline used
to compute layer-wise redundancy scores in Section~\ref{sec:activation_capture}.
All activation collection was performed on frozen models without gradient tracking.

\paragraph{Forward-pass instrumentation.}
For each decoder layer \( \ell \), we register a pre-forward hook to capture its
input hidden states \( H^{\mathrm{in}}_{\ell} \) and a post-forward hook to capture
its output hidden states \( H^{\mathrm{out}}_{\ell} \).
Both tensors are collected in floating-point precision and stored on CPU memory after each
forward step. No modifications are made to the model architecture or weights.

\paragraph{Tokenization and modality formatting.}
Each example consists of a single user message containing both an image and a
text prompt. Images are represented using the model's built-in visual token
format (vision start tokens, image patch tokens, padding tokens, and vision end
tokens). The text prompt is appended using the model's standard chat template,
including role tags and system tokens if required by the underlying model.

\paragraph{Pooling.}
To reduce storage and enable baselines to reuse activations, we compute mean-pooled
representations
\[
\bar{h}^{\mathrm{in}}_{\ell},\quad
\bar{h}^{\mathrm{out}}_{\ell} \in \mathbb{R}^{d}
\]
by averaging across both batch and token dimensions.
These vectors preserve coarse information about layer transformations and enable
methods such as CKA and Interlace to operate without access to full activations.

\paragraph{Domains and sample sizes.}
We collect activations separately for math and non-math domains.
The math domain uses 5{,}000 image--prompt examples, while the non-math
domain uses 4{,}000 examples, spanning captioning, entity listing, count-based
VQA, and grounding tasks. These datasets are described in
Section~\ref{sec:datasets}.

\subsection{Prompts Used for Math and Non-Math Probing}
\label{sec:prompts_formatted}

We use a fixed set of task prompts to ensure consistent activation
probing across math and non-math domains. Each prompt is paired with an
image and inserted into the model's multimodal chat template.

\subsubsection*{Math Domain Prompts}

\begin{tcolorbox}[title=Math-CoT (Step-by-Step Reasoning)]
Solve the following mathematical problem step by step. Explain your reasoning clearly
and provide the final answer.  
Problem: Look at the math problem in the image and solve it.
\end{tcolorbox}

\begin{tcolorbox}[title=Math-Direct (Answer Only)]
Solve the following mathematical problem. Provide only the final answer, without
explanation.  
Problem: Look at the math problem in the image.
\end{tcolorbox}

\begin{tcolorbox}[title=Math-Rephrase (Problem Paraphrasing)]
Rephrase the following math problem in simpler words, without solving it.  
Problem: Look at the math problem in the image.
\end{tcolorbox}

\begin{tcolorbox}[title=Math-Formalize (Equation Extraction)]
Convert the following math problem into a set of mathematical equations. Do not solve
the equations.  
Problem: Look at the math problem in the image.
\end{tcolorbox}

\begin{tcolorbox}[title=Math-Verify (Feasibility Assessment)]
A person claims they solved the math problem in the image. Look at the problem and
determine if it appears solvable. Respond with your assessment and explain briefly.
\end{tcolorbox}

\subsubsection*{Non-Math Prompts}

\begin{tcolorbox}[title=Captioning (Global Description)]
Describe the image in one complete, factual sentence. Avoid opinions or speculation.
\end{tcolorbox}

\begin{tcolorbox}[title=Entity Listing]
List the main objects or entities visible in the image. Return a comma-separated list
using short noun phrases.
\end{tcolorbox}

\begin{tcolorbox}[title=VQA: Counting]
Question: How many main objects are visible in the image?  
Answer with a single number.
\end{tcolorbox}

\begin{tcolorbox}[title=Grounding (Referring Expression)]
Identify the object referred to by the phrase ``the most prominent object'' and describe
it briefly using one short phrase.
\end{tcolorbox}

\subsection{Interpretation of Baseline and Baseline+SFT Results}

The baseline+SFT results do not represent an attempt to improve the base model's task performance, as shown in Table~\ref{tab:baseline_sft_reference}. Instead, this brief fine-tuning stage is designed solely to \emph{restore generation stability} after 
structural pruning. Because the SFT mixture is dominated by general multimodal instruction-following data 
rather than task-specific supervision (for example, ChartQA- or Snapask-style data), it can shift the 
model's behavior away from the original pre-training distribution. As a result, baseline+SFT sometimes 
underperforms the purely unpruned baseline on certain datasets. This effect is expected: the SFT stage 
is intended to ensure that \emph{pruned models remain functional}, not to serve as a performance booster 
for the unpruned checkpoints.

\begin{table}[h]
\centering
\footnotesize
\setlength{\tabcolsep}{4pt}
\renewcommand{\arraystretch}{1.0}
\caption{Reference accuracies for unpruned baselines. ``Baseline'' is the base model without fine-tuning. ``Baseline+SFT'' is the same model after the fixed SFT procedure used for post-pruning stabilization.}
\label{tab:baseline_sft_reference}
\begin{tabular}{lcccc}
\hline
\textbf{Dataset} & \textbf{2B Base} & \textbf{2B Base+SFT} & \textbf{4B Base} & \textbf{4B Base+SFT} \\
\hline
Snapask & 40.47 & 32.83 & 50.75 & 44.96 \\
NuminaMath & 63.45 & 52.61 & 73.49 & 64.06 \\
LLaVA-OneVision & 59.12 & 62.00 & 67.26 & 68.00 \\
ChartQA & 79.72 & 32.60 & 84.00 & 35.12 \\
RealWorldQA & 65.23 & 66.93 & 71.50 & 72.29 \\
ScienceQA & 86.47 & 87.80 & 92.71 & 93.46 \\
VStar & 78.01 & 78.01 & 81.68 & 82.20 \\
\hline
\end{tabular}
\end{table}

\clearpage
\subsection{Additional Figures}

\begin{figure*}[ht!]
\centering
\includegraphics[width=\textwidth]{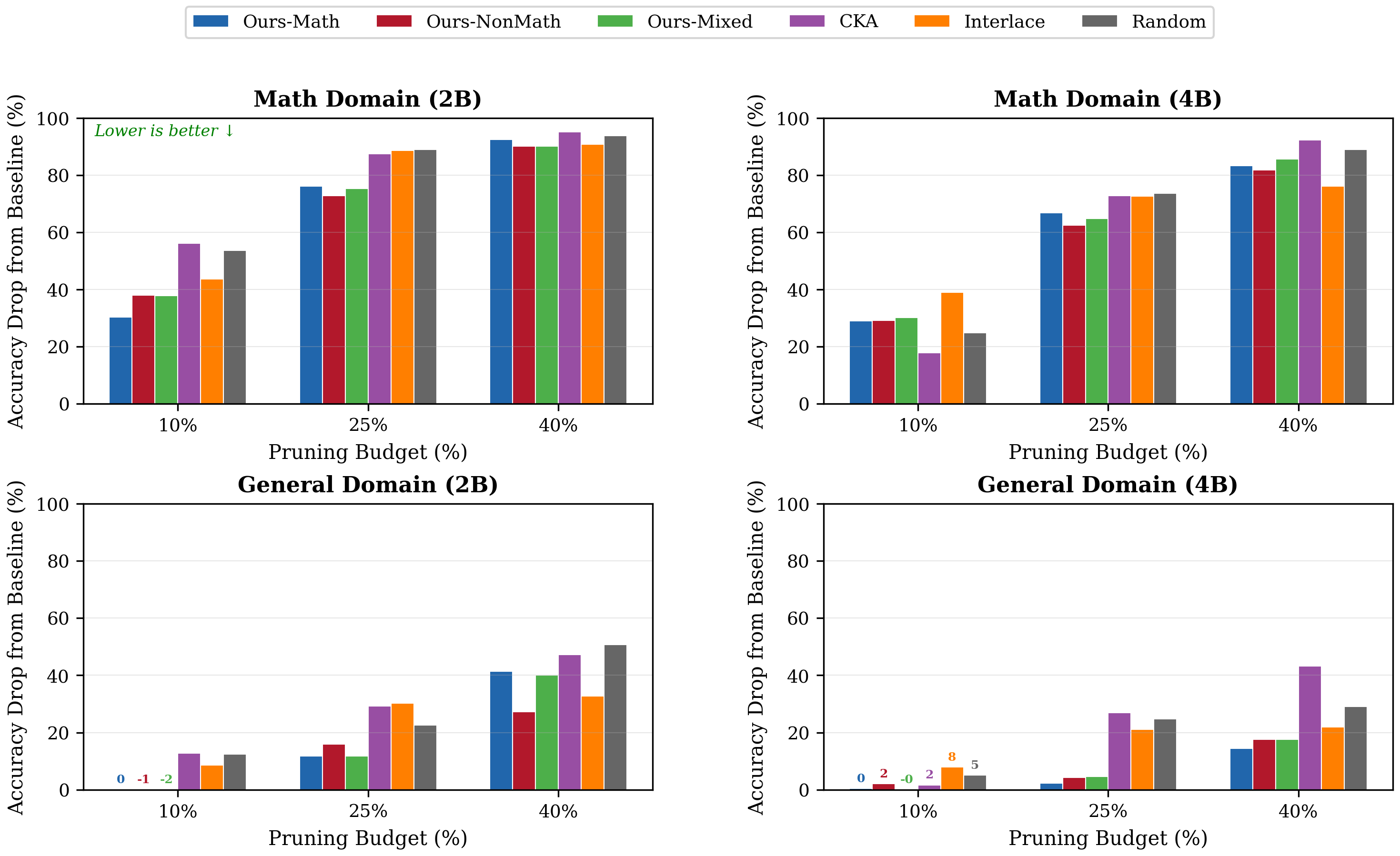}
\caption{Relative accuracy drop from baseline by domain, model size, and pruning budget.}
\label{fig:relative_drop_bars}
\end{figure*}

\begin{figure*}[ht!]
\centering
\includegraphics[width=\textwidth]{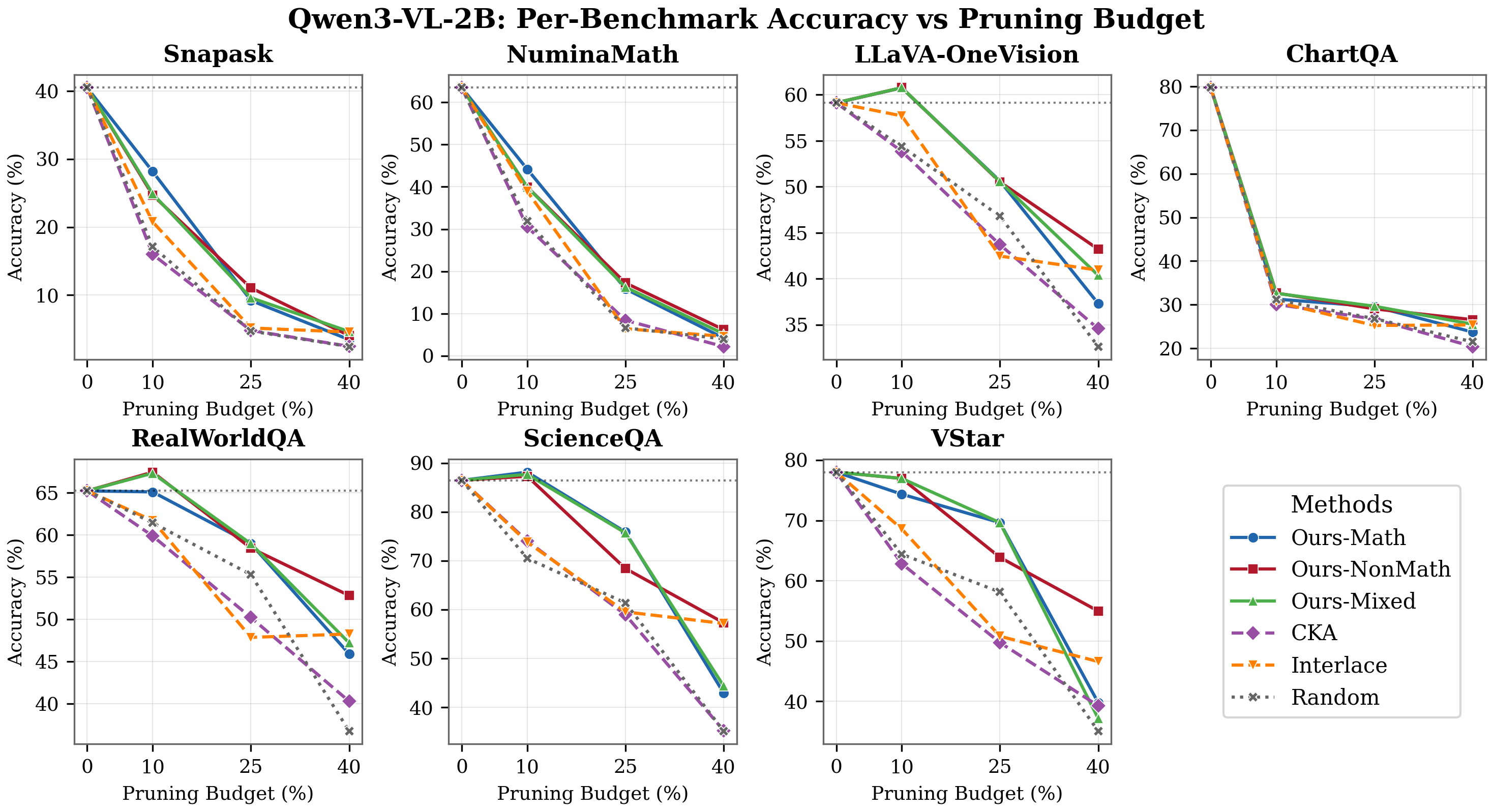}
\caption{Per-benchmark accuracy curves for Qwen3-VL-2B.}
\label{fig:per_benchmark_curves_2b}
\end{figure*}

\begin{figure*}[ht!]
\centering
\includegraphics[width=\textwidth]{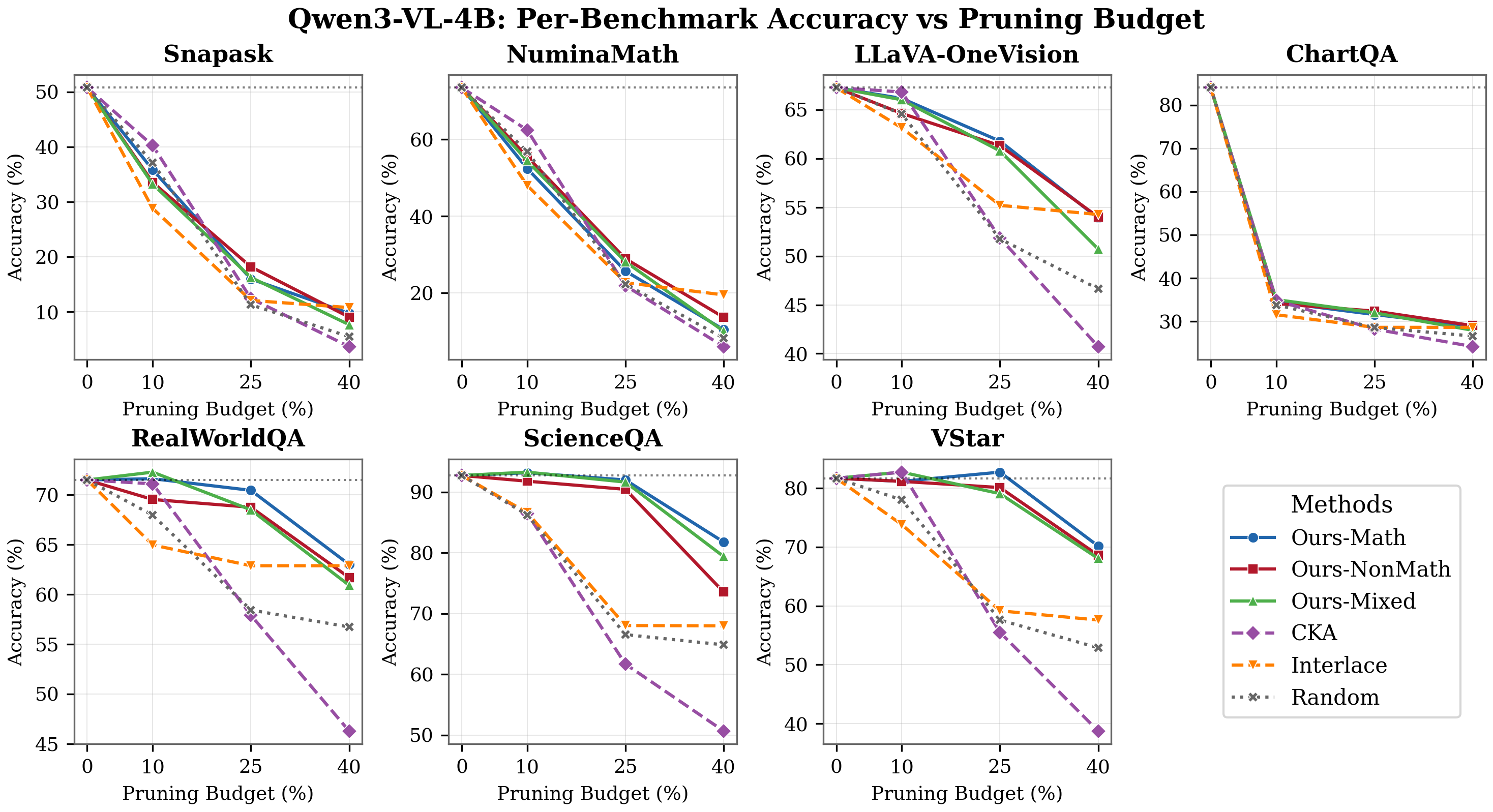}
\caption{Per-benchmark accuracy curves for Qwen3-VL-4B.}
\label{fig:per_benchmark_curves_4b}
\end{figure*}

\begin{figure*}[ht!]
\centering
\includegraphics[width=\textwidth]{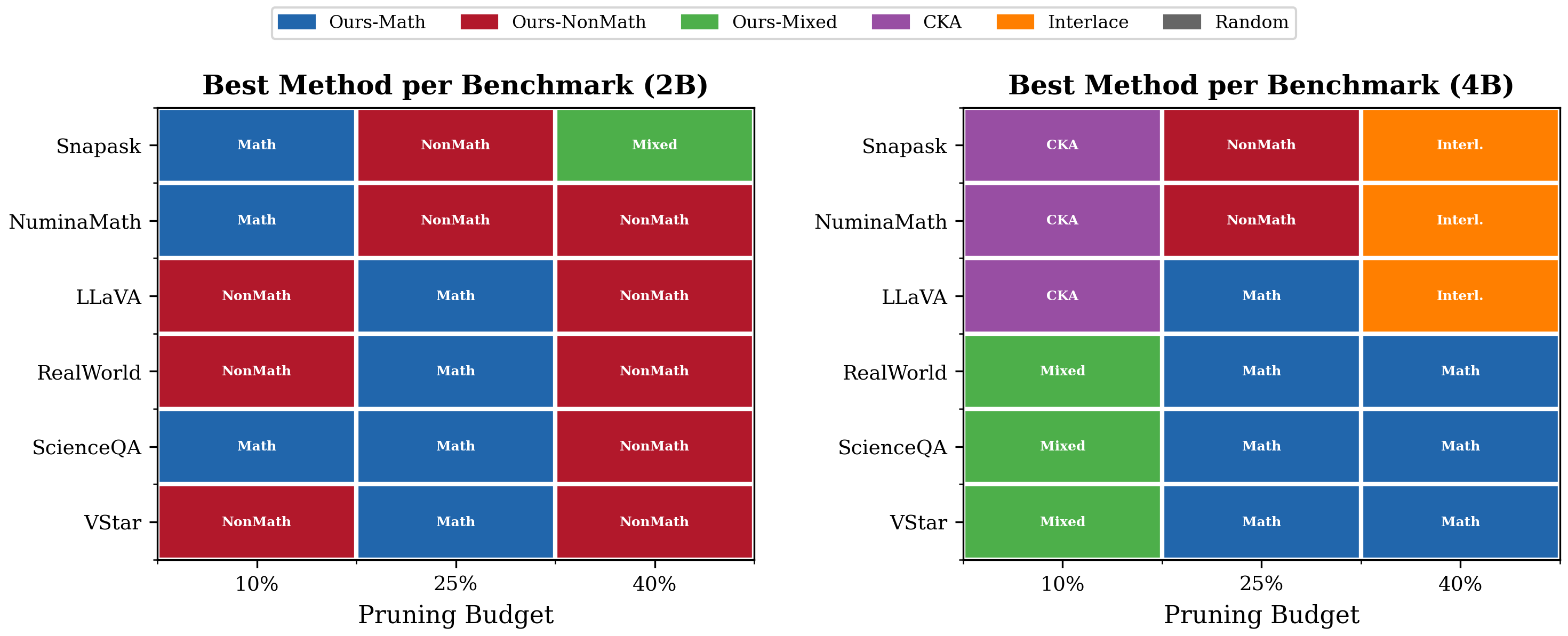}
\caption{Best method per benchmark and pruning budget for 2B and 4B.}
\label{fig:winner_heatmap}
\end{figure*}

\end{document}